\documentclass[conference]{IEEEtran}
\IEEEoverridecommandlockouts
\usepackage{cite}
\usepackage{hyperref}
\usepackage{amsmath,amssymb,amsfonts}
\usepackage{graphicx}
\usepackage{textcomp}
\usepackage{subcaption}
\usepackage{tikz}

\newcommand\copyrighttext{%
  \footnotesize \textcopyright 2024 IEEE.  Personal use of this material is permitted.  Permission from IEEE must be obtained for all other uses, in any current or future media, including reprinting/republishing this material for advertising or promotional purposes, creating new collective works, for resale or redistribution to servers or lists, or reuse of any copyrighted component of this work in other works.}
\newcommand\copyrightnotice{%
\begin{tikzpicture}[remember picture,overlay]
\node[anchor=south,yshift=10pt] at (current page.south) {\fbox{\parbox{\dimexpr\textwidth-\fboxsep-\fboxrule\relax}{\copyrighttext}}};
\end{tikzpicture}%
}

\def\BibTeX{{\rm B\kern-.05em{\sc i\kern-.025em b}\kern-.08em
    T\kern-.1667em\lower.7ex\hbox{E}\kern-.125emX}}
\begin{document}

\title{Iterative Filter Pruning for Concatenation-based CNN Architectures}

\author{\IEEEauthorblockN{Svetlana Pavlitska$^{1,2}$, 
Oliver Bagge$^{2}$, 
Federico Peccia$^{1,3}$, 
Toghrul Mammadov$^{2}$, 
and J.~Marius Zöllner$^{1,2}$}
\IEEEauthorblockA{
\textit{$^{1}$ FZI Research Center for Information Technology} \\
\textit{$^{2}$ Karlsruhe Institute of Technology (KIT)}\\
\textit{$^{3}$ University of Tübingen, Tübingen, Germany}\\
Karlsruhe, Germany \\
pavlitska@fzi.de}
}

% \author{\IEEEauthorblockN{1\textsuperscript{st} Given Name Surname}
% \IEEEauthorblockA{\textit{dept. name of organization (of Aff.)} \\
% \textit{name of organization (of Aff.)}\\
% City, Country \\
% email address or ORCID}
% \and
% \IEEEauthorblockN{2\textsuperscript{nd} Given Name Surname}
% \IEEEauthorblockA{\textit{dept. name of organization (of Aff.)} \\
% \textit{name of organization (of Aff.)}\\
% City, Country \\
% email address or ORCID}
% \and
% \IEEEauthorblockN{3\textsuperscript{rd} Given Name Surname}
% \IEEEauthorblockA{\textit{dept. name of organization (of Aff.)} \\
% \textit{name of organization (of Aff.)}\\
% City, Country \\
% email address or ORCID}
% \and
% \IEEEauthorblockN{4\textsuperscript{th} Given Name Surname}
% \IEEEauthorblockA{\textit{dept. name of organization (of Aff.)} \\
% \textit{name of organization (of Aff.)}\\
% City, Country \\
% email address or ORCID}
% \and
% \IEEEauthorblockN{5\textsuperscript{th} Given Name Surname}
% \IEEEauthorblockA{\textit{dept. name of organization (of Aff.)} \\
% \textit{name of organization (of Aff.)}\\
% City, Country \\
% email address or ORCID}
% \and
% \IEEEauthorblockN{6\textsuperscript{th} Given Name Surname}
% \IEEEauthorblockA{\textit{dept. name of organization (of Aff.)} \\
% \textit{name of organization (of Aff.)}\\
% City, Country \\
% email address or ORCID}
% }

\maketitle

\copyrightnotice
\thispagestyle{empty}
\pagestyle{empty}

\begin{abstract}
Model compression and hardware acceleration are essential for the resource-efficient deployment of deep neural networks. Modern object detectors have highly interconnected convolutional layers with concatenations. In this work, we study how pruning can be applied to such architectures, exemplary for YOLOv7. We propose a method to handle concatenation layers, based on the connectivity graph of convolutional layers. By automating iterative sensitivity analysis, pruning, and subsequent model fine-tuning, we can significantly reduce model size both in terms of the number of parameters and FLOPs, while keeping comparable model accuracy. Finally, we deploy pruned models to FPGA and NVIDIA Jetson Xavier AGX.  Pruned models demonstrate a 2x speedup for the convolutional layers in comparison to the unpruned counterparts and reach real-time capability with 14 FPS on FPGA. Our code is available at \url{https://github.com/fzi-forschungszentrum-informatik/iterative-yolo-pruning}.
\end{abstract}


\section{Introduction}
Computer vision tasks rely heavily on convolutional neural networks (CNNs). CNN architectures have become increasingly deep and complex with millions of parameters, making their deployment for energy-efficient tasks challenging~\cite {fleck2023low}. On the other hand, CNNs are known to be heavily overparameterized~\cite{frankle2018lottery,pavlitska2023relationship} and thus possess a large potential for compression.

Pruning~\cite{han2015deep} is one of the established techniques to reduce model size. Unstructured pruning, while simple to implement, does not provide any benefits in model size. To obtain smaller and faster networks, structured pruning needs to be applied which requires architectural modifications. Simple networks such as VGG-16~\cite{simonyan2014very} comprise a single stack of convolutional layers, making them relatively easy to prune structurally. If a convolutional layer is pruned, only the layer immediately following it is affected. Structured filter pruning has been successfully applied to VGG and ResNet models~\cite{li2017pruning}.

The architecture of modern CNNs, especially for tasks like object detection, is much more complex. For example, YOLOv7~\cite{wang2022yolov7} has the skip connections and efficient layer aggregation network (ELAN) modules~\cite{wang2023designing} with concatenation layers, which results in extensive branching. Concatenation layers combine the output feature maps of several convolutional layers along the channel dimension to produce a larger feature map that serves as an input for subsequent layers. It is challenging to identify which layers get affected when pruning a convolutional layer. YOLOv8~\cite{jocher2023yolov8} even employs both concatenation layers and residual connections.

\begin{figure}[t]
  \centering
  \includegraphics[width=\columnwidth]{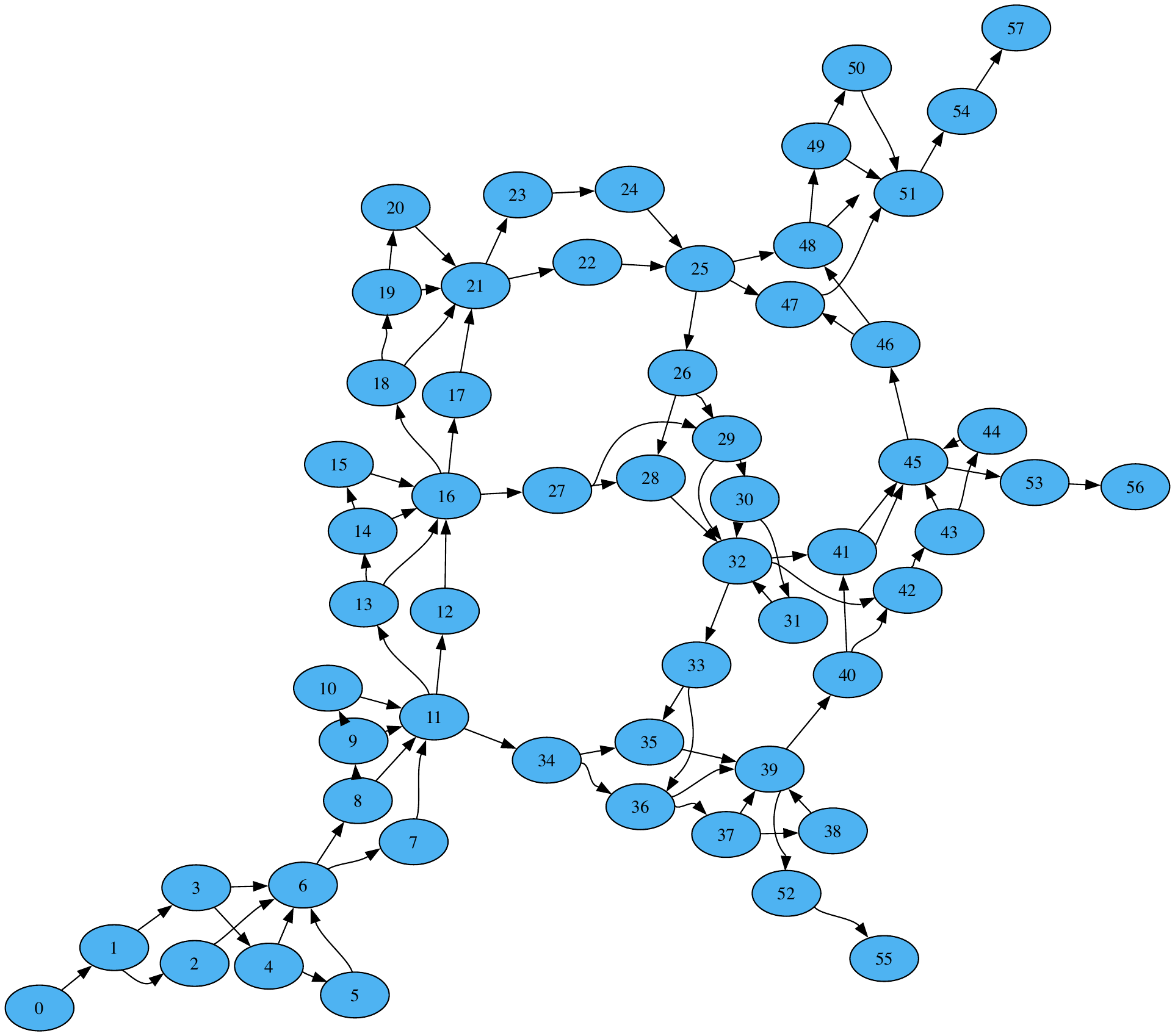}
  \caption{The proposed iterative pruning approach relies on a layer connectivity graph (here exemplary for YOLOv7-tiny).}
  \label{fig:yolov7-tiny-graph}
\end{figure}

This structural complexity of concatenation layers makes pruning especially challenging. Some recent works even avoid concatenation layers during pruning~\cite{zhang2021pruned}. In this work, we propose to use a connectivity graph as an intermediate model representation (see Figure~\ref{fig:yolov7-tiny-graph}). For each convolutional layer, it represents all connections to subsequent layers, which should be considered during pruning. Concatenation layers are thus implicitly incorporated in pruning. Furthermore, sensitivity analysis and automated selection of layers and pruning rates help to iteratively perform pruning even for complex architectures, choosing new pruning parameters at each iteration. 

We evaluate the described method on two variants of a concatenation-based object detector YOLOv7. We analyze the impact of the pruning strategy on each layer in terms of the number of parameters and FLOPs. Finally, we evaluate pruned models on two different edge AI platforms and assess the impact of pruning on the model acceleration pipeline. 

In this work, we demonstrate, how the proposed method helps to gain up to 80\% sparsity for YOLOv7 and YOLOv7-tiny models without a considerable drop in detection accuracy and achieve a 2x speedup for the pruned convolutional layers as well as a real-time inference speed of 14 FPS on FPGA.

\section{Related Work}

%\subsection{Pruning Convolutional Neural Networks}
%\label{sec:pruning_cnn}
%Unstructured pruning for CNNs is similar to that in fully connected networks. Single weights in the layer's filters are set to zero in each convolutional layer. Here, too, we do not gain anything from this because of the relatively low sparsity and the size and number of filters staying the same.

Deep learning models are usually over-parameterized and possess a certain redundancy~\cite{zhang2017understanding}, which can be reduced to save both computation and memory resources. A plethora of methods aiming at the reduction of memory requirements for deep learning models have been proposed so far, whereas the most popular groups of approaches are pruning and quantization.
%including model distillation~\cite{hinton2015distilling}, neural architecture search 

Pruning techniques can be grouped into \textit{unstructured} methods, which aim at removing connections~\cite{han2016eie} and \textit{structured pruning} approaches, aiming at deleting certain structures from a network, e.g., neurons from fully connected layers, filters from convolutional layers, or self-attention heads from transformers. Examples of unstructured pruning methods are early works on Optimal Brain Damage~\cite{le2021network} and Optimal Brain Surgeon~\cite{lecun1989optimal}, and magnitude-based pruning. %Han et al.~\cite{han2017dsd} iterate training, magnitude-based pruning and fine-tuning.

\subsection{Filter Pruning for CNNs}
In this work, we focus on structured pruning for CNNs, predominantly presented by filter pruning.
Li et al.~\cite{li2017pruning} explored one-shot structured pruning on VGG-16~\cite{simonyan2014very} and ResNet~\cite{he2016deep} models that were pre-trained on the CIFAR-10~\cite{Krizhevsky09learningmultiple} image classification dataset. Specific layers are pruned by removing the filters with the smallest L1-norm. When removing filters in a convolutional layer, the kernels corresponding to the missing feature map are removed in the next convolutional layer. Hereby, residual blocks need special attention because pruning a layer can affect multiple other layers. Batch normalization layers after the pruned layer also have to be modified. When pruning multiple layers at once, filters can be selected independently or greedily. The independent approach selects filters without considering the effects of the pruning of different layers on the current layer. In the greedy approach, pruned kernels are not included in calculating the pruning criterion. Greedy pruning results in networks with higher accuracy. They show that pruning filters with the smallest L1-norm is better than pruning random filters or those with the largest L1-norm, but no noticeable difference between using the L1-norm and L2-norm was found. 

To find out which layers can be pruned, a sensitivity analysis is conducted. By measuring the accuracy drop when a single layer is pruned on its own, it can be assessed which layers are more sensitive to pruning than others. More sensitive layers generally perform worse across pruning rates than less sensitive ones.  Also, earlier layers are more sensitive to pruning than deeper layers. Based on these results, layers are selected manually for pruning.

%\subsection{Fine-Tuning a Pruned Network}
%\label{sec:retraining_related}
Neural networks can be re-trained to regain the loss in accuracy caused by pruning. The simplest way to do this is to use the same smallest learning rate at the end of training the original model (fine-tuning). This method was employed by Li et al.~\cite{li2017pruning} by using a constant learning rate of 0.001 and fine-tuning for 40 epochs. They could prune 64\% of parameters in VGG-16 and 32.4\% of parameters in ResNet-110 while reducing the number of FLOPs by 34.2\% in VGG-16 and 38.6\% in ResNet-110. Fine-tuning almost fully restored the accuracy of the unpruned model. They find that training a pruned model from scratch leads to worse accuracies than pruning a pre-trained model and fine-tuning it afterward.

Renda et al.~\cite{renda2020comparing} and Le et al.~\cite{le2021network} compared fine-tuning with a small, constant learning rate to using learning rate schedules with larger learning rates. %Learning rate rewinding, for example, rewinds the learning rate to the value it had in an earlier training epoch or even to the start of training. 
When fine-tuning with a small number of epochs, they find that using a small, constant learning rate converges faster than other methods. Other learning rate schedules lead to higher accuracies but come with a significant increase in training time.

%\subsection{Iterative Pruning}
%\label{sec:iterative_pruning}
Instead of pruning and fine-tuning once, it is possible to do this iteratively. An important factor that enables this is sparsity training~\cite{wen2016learning}. Sparsity training applies regularization techniques during training that encourage certain parameters to be pushed toward zero. This is done by adding regularization terms to the loss function, such as the L1 or L2 norm of weights~\cite{NIPS2015_ae0eb3ee}. After sparsity training, parameters close to zero can be removed, and the sparsity training can be repeated.

\subsection{Pruning of Object Detectors}

SlimYOLOv3 approach by Zhang et al.~\cite{zhang2019slimyolo}) applies channel pruning as proposed by Li et al.~\cite{liu2017learning} to YOLOv3~\cite{redmon2018yolov3}. Channels are selected for pruning that correspond to the smallest scale factors in the subsequent batch normalization layer. Sparsity training is applied by imposing L1-regularization on these scale factors during training. YOLOv3 is trained from scratch with sparsity training for 100 epochs. After pruning, the model is fine-tuned with the same hyper-parameters as YOLOv3 to compensate for the degradation in accuracy. As a result, the authors reduced FLOPs by 90.8\%, and the number of parameters by 92\%, while achieving comparable detection accuracy on the VisDrone2018-Det dataset~\cite{zhuvisdrone2018}. 

A follow-up approach called Pruned-YOLO~\cite{zhang2021pruned} improves on SlimYOLOv3 by combining the fine-tuning and retraining steps. The approach is applied to YOLOv3 and YOLOv5 on the VisDrone2018-Det~\cite{zhuvisdrone2018} and COCO~\cite{lin2015microsoft}. The authors also change the criterion for selecting unimportant filters and also consider the weights of the convolutional layers. %Their importance measurement $\theta_i$ for a channel is given by Equation~\ref{eqn:pruned_yolo_importance}. $C_{out}$ is the number of channels in the output feature map equivalent to the number of filters in the layer. $\alpha_i$ represents the $i$'th filter in the convolutional layer. $\gamma_i$ is the scale factor of the batch normalization layer following the convolutional layer corresponding to the $i$'th channel.

% \begin{equation}
% \theta_i = |\gamma_i| * \sum\limits_{t \in \alpha_i} |t|, i \in [1, 2, ..., C_{out}]
% \label{eqn:pruned_yolo_importance}
% \end{equation}

% The absolute value of the channel's scale factor in the batch normalization layer is multiplied by the L1-norm of the filter weights that produce that channel.
%Iterative pruning may yield better results but takes much more training epochs. One-shot pruning can remove many filters and regain accuracy from fine-tuning for fewer epochs.

%\textbf{TODO}: what level of sparsity is reached in other works, at the cost of what drop in mAP,  Existing works, explicitly applying iterative filter pruning to complex CNN architectures, ....  To perform pruning using existing frameworks (e.g. built-in PyTorch pruning), a user is expected to define what layers and what pruning rate to take.

Ahn et al.~\cite{ahn_safp-yolo_2023} pruned YOLOv4 and YOLOV7 models using spatial attention-based filter pruning on pig and vehicle datasets. With the pruning rate of 87.5\%, they achieved 72.21\%, 93.33\%, 70.1\%, and 93.44\% parameters drop for YOLOv7, YOLOv7-tiny, YOLOv4 and YOLOv4-tiny respectively with at most 8.79\% AP@50 drop. By sacrificing 22.75\% mAP, Li et al.~\cite{li_channel_2022} reached 96.02\% and 95.08\% sparsity levels for parameters and FLOPS respectively using the lossless reconstruction of the layers of YOLOv3 with a 90\% pruning rate. %Instead of using the fixed pruning rate for all convolutional layers, Yang and Liu~\cite{yang_channel_2022} determined a pruning rate for each layer of ResNet56 based on the layer sensitivity and were able to achieve a 74.35\% parameters drop as well as a 72.23\% FLOPS drop with only 2.12\% accuracy drop on CIFAR10. On the other hand, Wang et al.~\cite{wang2021convolutional} reached 53.8\% sparsity on FLOPS by pruning ResNet56 using structural redundancy reduction with graph redundancy and achieved 0.37\% increase on accuracy.

The complex architecture of further YOLO versions, starting from YOLOv6, and especially strong interconnections due to a large number of concatenation layers make pruning challenging~\cite{jani2023model}. Furthermore, works applying iterative filter pruning to the latest YOLO versions are sparse. %Excplicit description of pruning strategies can be found only for YOLOv3 in~\cite{zhang2019slimyolo} and for YOLOv5 in ~\cite{zhang2021pruned}. Furthermore, we could not find any works explicitly handling concatenation layers.
Zhang et al.~\cite{zhang2023channel} apply one-shot pruning to downsize the YOLOv7 model from the original 37M to 25M parameters, thus reaching the sparsity level of 32\%, and perform the runtime evaluation on GPU. Furthermore, open-source applications that we have found, focus on small sparsity rates $<0.1$. The only work explicitly handling the concatenation layer is that by Fang et al.~\cite{fang2023depgraph}, where a library for structured pruning is proposed. They also use a graph-based approach, but in addition to concatenation, also consider other types of dependency: residual and reduction. The authors, however, do not analyze the gain of sparsity for the inference speedup.

In summary, the pruning of concatenation-based architectures with high sparsity levels is rarely explicitly addressed in literature. Furthermore, the implications of high sparsity on inference speed are either completely overlooked or analyzed on only one type of hardware.

\section{Method}

We propose an automated iterative pruning method targeting concatenation-based architectures. It relies on a connectivity graph to automatically prune all affected layers, including concatenations. It iteratively conducts sensitivity analysis to identify layers that retain high accuracy after the deletion of a certain portion of parameters or FLOPs from them. % To the best of our knowledge, structured filter pruning has not yet been applied to a network with concatenation layers.

\subsection{Handling Concatenation Layers}
\label{sec:pruning_with_concat_layers}

\begin{figure}[t]
  \centering
  \includegraphics[width=0.7\columnwidth]{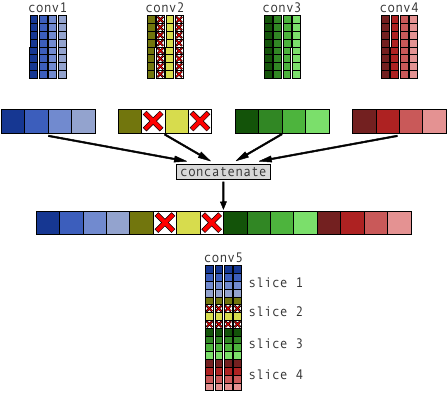}
  \caption{Concatenation layers during pruning: two filters of the \texttt{conv2} layer are pruned, so the kernels corresponding to filters should be removed from the input feature map of the \texttt{conv5}.}
  \label{fig:concatenation_example}
\end{figure}

For structured pruning, the only types of layers that need to be modified are the convolutional layers. Concatenation, max-pooling, and upsampling layers are not affected by filter pruning. However, for concatenation-based architectures, concatenation layers should be explicitly handled during pruning. As Figure~\ref{fig:concatenation_example} shows, missing kernels in the input feature map of the concatenation layer should be dropped.

Furthermore, concatenation layers can output their result to multiple consecutive layers. If a concatenation layer passes its output first to a max-pool or upsampling layer, the convolutional layer after the max-pool or upsampling layer needs to be modified. To automate the described process, we propose an intermediate network representation called the \textit{layer connectivity graph}. It allows us to prune convolutional layers by providing a layer index and a pruning rate. The affected consecutive layers are then pruned automatically. %This will make it easier to try out different combinations of layers to prune later.

%The following section will look at how this process can be automated with the example of YOLOv7 so that these architectural changes are not made manually.

\subsection{Layer Connectivity Graph}
\label{sec:yolov7_config_file}

We define the \textit{connectivity graph} for a deep neural network $N$ as a directed weighted graph $G_N = (V, E, W)$ with a set of vertices $V=\{{v | v \in Conv(N)\}}$, where each vertex $v$ represents a layer in a set $Conv(v)$ of convolutional layers. 
% Each vertex stores a tuple of two values: 
% \begin{enumerate}
% %\item List of all the next layers affected by pruning the layer at index $i$. 
% \item $slice\_idx$ is a list of concatenation slices from which kernels in the filters of the next layers must be removed. This parameter is 0 if the output is not passed to a concatenation layer.
% \item $slice\_size$: the list of sizes of concatenation slices if a layer $v$ 
% \end{enumerate}
For a set of edges $E$, a pair of vertices $(x,y)$ is connected with an edge if the input of layer $x$ is used in a feature map of a layer $y$. A weight $w$ of an edge $e=(x,y) \in E$ represents a concatenation slice from which kernels in the filters of a layer $y$ must be removed when layer $x$ is pruned. $w$ is 0 if the output of $x$ is not passed to a concatenation layer. Finally, if a layer $x$ receives inputs from a concatenation layer, we additionally store the list of sizes of concatenation slices in the vertex $x$. %Figure~\ref{fig:conv-graph} demonstrates the connectivity graph exemplary for an ELAN module. 

\begin{figure}[t]
\centering
  \begin{subfigure}[t]{0.54\columnwidth}
    \centering
    \includegraphics[width=\textwidth]{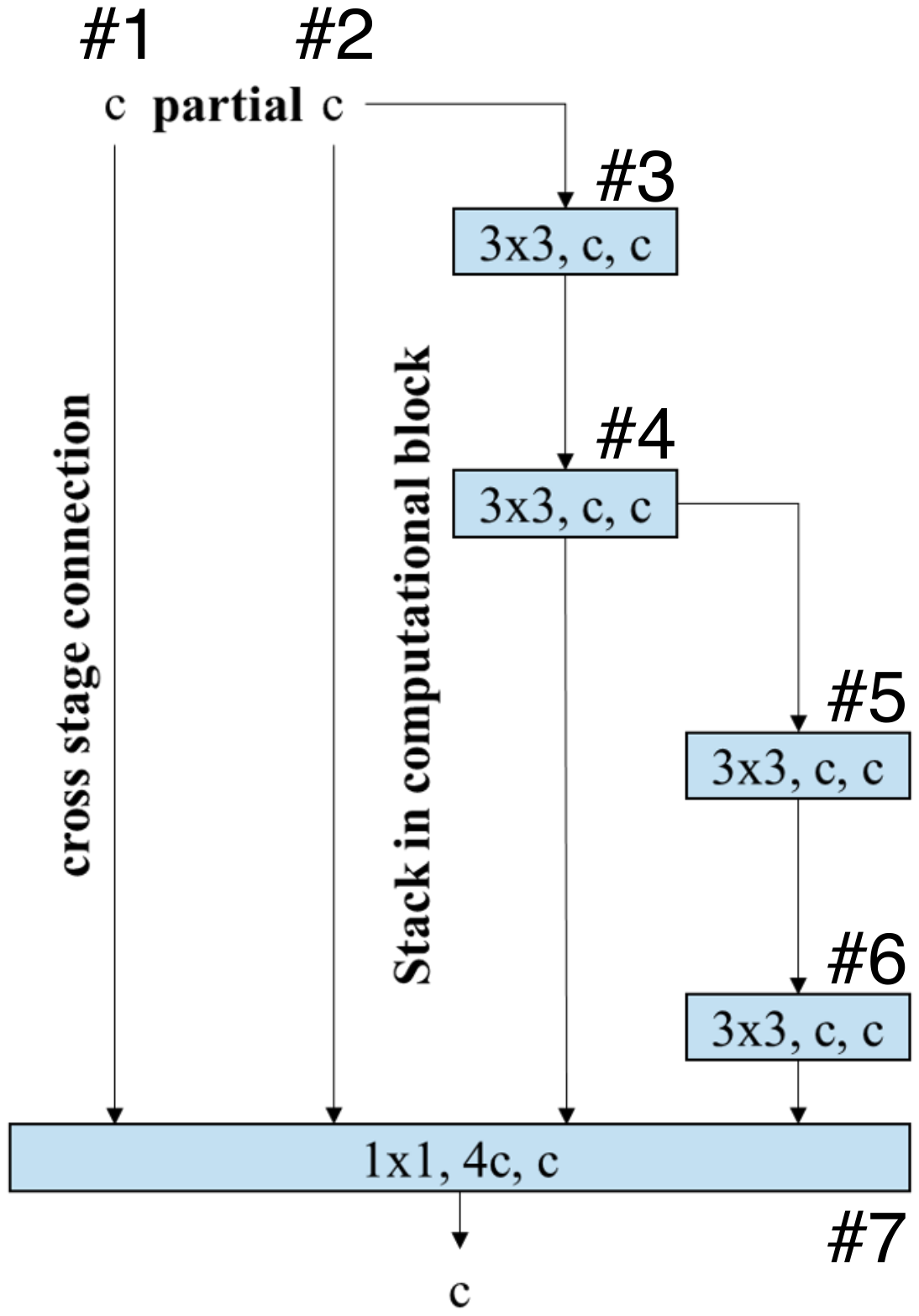}
    \caption{ELAN module~\cite{wang2023designing}}
  \end{subfigure}
  \begin{subfigure}[t]{0.41\columnwidth}
    \centering
    \includegraphics[width=\textwidth]{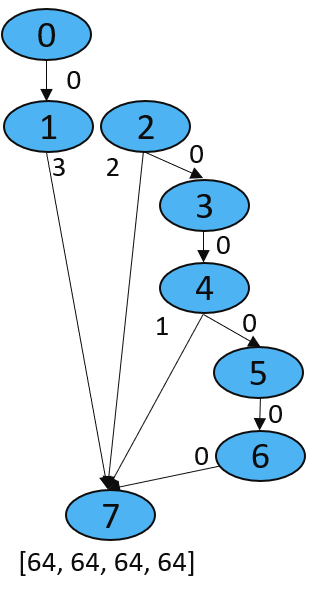}
    \caption{Connectivity graph}
  \end{subfigure}
  \caption{Connectivity graph of an ELAN module.}
  \label{fig:conv-graph}
\end{figure}

% \begin{figure*}[t]
%   \centering
%   \includegraphics[width=0.9\textwidth]{fig/conv_graph.png}
%   \caption{Example of a connectivity graph for convolutional and concatenation layers.}
%   \label{fig:connectivity_graph}
% \end{figure*}

To apply pruning to a layer $x$, we first delete filters from layer $x$. We then obtain a list of all layers $y$ affected by this pruning as $\{y | e=(x,y) \in E\}$ and then delete the corresponding filters from a concatenation slice $w(e)$ from a layer $y$. If the slice index $w(e)$ is greater than zero, the starting index of the correct slice needs to be determined with the help of the slice sizes stored in the vertex $y$.

For an ELAN module (see Figure~\ref{fig:conv-graph}), to prune layer $4$, the first kernels in this layer are removed by pruning along the first dimension of the convolutional layer’s weights. Next, kernels from layer $5$ should be deleted, now along the second dimension of the convolutional layer’s weights. Because the weight of an edge between $4$ and $5$ (i.e. slice index) is 0, the same kernel indices can be used. There is also an edge connecting $4$ to $7$ with a weight of 1. Since indexing begins with 0, kernels from the second slice in layer $7$ should be removed. We use the slice size stored in vertex $7$ for the second slice (i.e. 64) to update filter indices and thus ensure that kernels from the correct slice are removed.

In summary, although the connectivity graph contains only nodes for convolutional layers, it implicitly takes into consideration concatenation layers, thus allowing automated pruning of complex network architectures.

\subsection{Iterative Pruning}
\label{sec:selecting_layers_to_prune}

Using the described connectivity graph, we can automatically prune layers in an architecture. To determine which layers should be pruned at a given iteration and with what pruning rate, we apply sensitivity analysis. Similarly to Li et al.~\cite{li2017pruning}, we conduct a sensitivity analysis by pruning each layer individually following the algorithm described above and evaluating the model mAP on the validation set.  Li et al~\cite{li2017pruning} select filters entirely based on the results of their sensitivity analysis. We additionally consider the number of parameters and FLOPs per layer when selecting layers to prune.

Furthermore, we dynamically determine the most suitable pruning rate. For this, we conduct sensitivity analysis by pruning the layers for different pruning rates to analyze the impact of the pruning on the mAP, parameters, and FLOPs. The layers and the pruning rates are determined by the function $V_{l, r}(x, y) = x \cdot y^{a}$, where $x$ is either FLOPs or parameters pruned in a layer $l$ for pruning rate $r$, $y$ is mAP after pruning, and $a$ is a hyperparameter controlling the impact of mAP on the function.  If the value returned by this function exceeds some predefined threshold, the layer $l$ is selected for pruning. After pruning, fine-tuning is performed to restore the drop in accuracy. If mAP is greater than 0.3, we fine-tune for 15 epochs, otherwise for 20 epochs. If mAP is less than 0.1, we fine-tuned for 25 epochs. Because mAP for YOLOv7-tiny was significantly lower, we fine-tuned for 25 epochs.
%The function evaluates the influence of the pruning rates for each layer by multiplying the layers and pruning rates with high values are selected for pruning for FLOPs and parameters.

%After evaluating the models by pruning the selected layers with the chosen pruning rates, the models are fine-tuned for 15 to 25 epochs based on the mAP degradation as a result of pruning.%A fine-tuned model can either be pruned further or deployed.  

\section{Experiments}

\subsection{Experimental Setup}
We evaluate different variants of the \texttt{YOLOv7}~\cite{wang2022yolov7} architecture, using the official PyTorch implementation\footnote{https://github.com/WongKinYiu/yolov7}. We use the COCO dataset~\cite{lin2015microsoft} for evaluation and fine-tuning, the images are resized to $640\times640$. We train and evaluate models using NVIDIA GTX 1080Ti and NVIDIA RTX 2080Ti and further deploy them to edge devices. Similar to Wang et al.~\cite{wang2022yolov7}, we report AP @0.5:0.95.

The connectivity graph for \texttt{YOLOv7} contains 91 nodes, and for \texttt{YOLOv7-tiny} 57 nodes (see Figure~\ref{fig:yolov7-tiny-graph}. We apply the proposed method to all nodes in the connectivity graph, omitting only the three last layers in the detection head.

We have performed 14 iterations of the proposed automated pruning on \texttt{YOLOv7} and \texttt{YOLOv7-tiny}. To obtain a realistic assessment of the performance of the resulting networks, we have deployed them to two different edge architectures.

\subsection{Ablation Studies}
\subsubsection{Pruning Before vs. After Fusing Batch Normalization Layers}
\label{subsec:pruning_before_or_after_bn}
\begin{figure}[t]
  \centering
  \includegraphics[width=0.7\columnwidth]{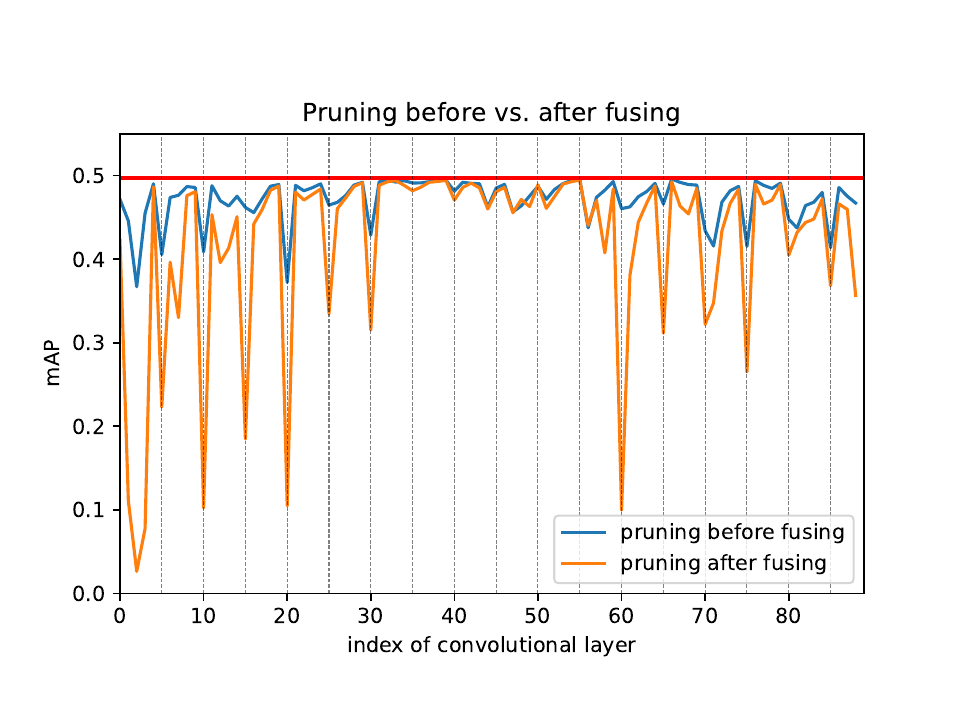}
  \caption{Pruning 50\% of filters in each layer based on the smallest L2-norm before (blue) vs. after (orange) fusing the batch normalization layers into the convolutional layers.} %The horizontal red line is the mAP of YOLOv7 (0.497).}
  \label{fig:before_vs_after_fusing}
\end{figure}

Batch normalization~\cite{ioffe2015batch} reduces the internal covariate shift by normalizing activations between layers for each mini-batch during training. It is common practice to fuse batch normalization layers into their preceding convolutional layers before inference to reduce the number of computations~\cite{mao2021neuralnetwork}. When pruning filters after this fusion, only the convolutional layers resulting from the fusion need to be modified. When pruning before the fusion, the parameters of the convolutional layers and the batch normalization layer need to be removed.
%To compare both variants, we conducted a sensitivity analysis by pruning every layer individually and measuring the resulting mAP on the validation set. 
%In Figure~\ref{fig:before_vs_after_fusing}, we compare the sensitivity of pruning individual layers before and after the fusion. The blue curve represents the mAPs for pruning before fusing, and the orange curve represents the mAPs for pruning after fusing. 50\% of filters with the smallest L2-norm are pruned for each layer. The red horizontal line corresponds to the mAP of the unpruned YOLOv7 model (0.497).
Our evaluation (see Figure~\ref{fig:before_vs_after_fusing}) shows that pruning filters after fusing the batch normalization layer has a significantly worse effect on the mAP than pruning before fusing.% To additionally quantify the results of the sensitivity analysis, we can calculate the average of the mAP values over all layers. %For pruning before fusing, the average mAP is 0.472, and for pruning after fusing it is 0.421. %The reason why pruning after fusing performs worse can be explained by the findings in Section~\ref{subsec:filter_importance_criteria}.

\begin{table}[h]
\centering
\begin{tabular}{|r|l|}
\hline
\textbf{Criterion} & \textbf{Average mAP} \\
\hline
Smallest L2 & \textbf{0.472} \\
Smallest L1 & 0.463 \\
Random & 0.454 \\
Smallest L1 * bn scaling factor~\cite{zhang2021pruned} & 0.451 \\
bn scaling factor~\cite{liu2017learning} & 0.443 \\
Largest L2 & 0.429 \\
\hline
\end{tabular}
\caption{Average mAP for different criteria (see Figure~\ref{fig:filter_criteria_sensitivity})}
\label{tab:average_mAP}
\end{table}

\subsubsection{Filter Importance Criteria}
\label{subsec:filter_importance_criteria}
\begin{figure}[t]
  \centering
  \includegraphics[width=\columnwidth]{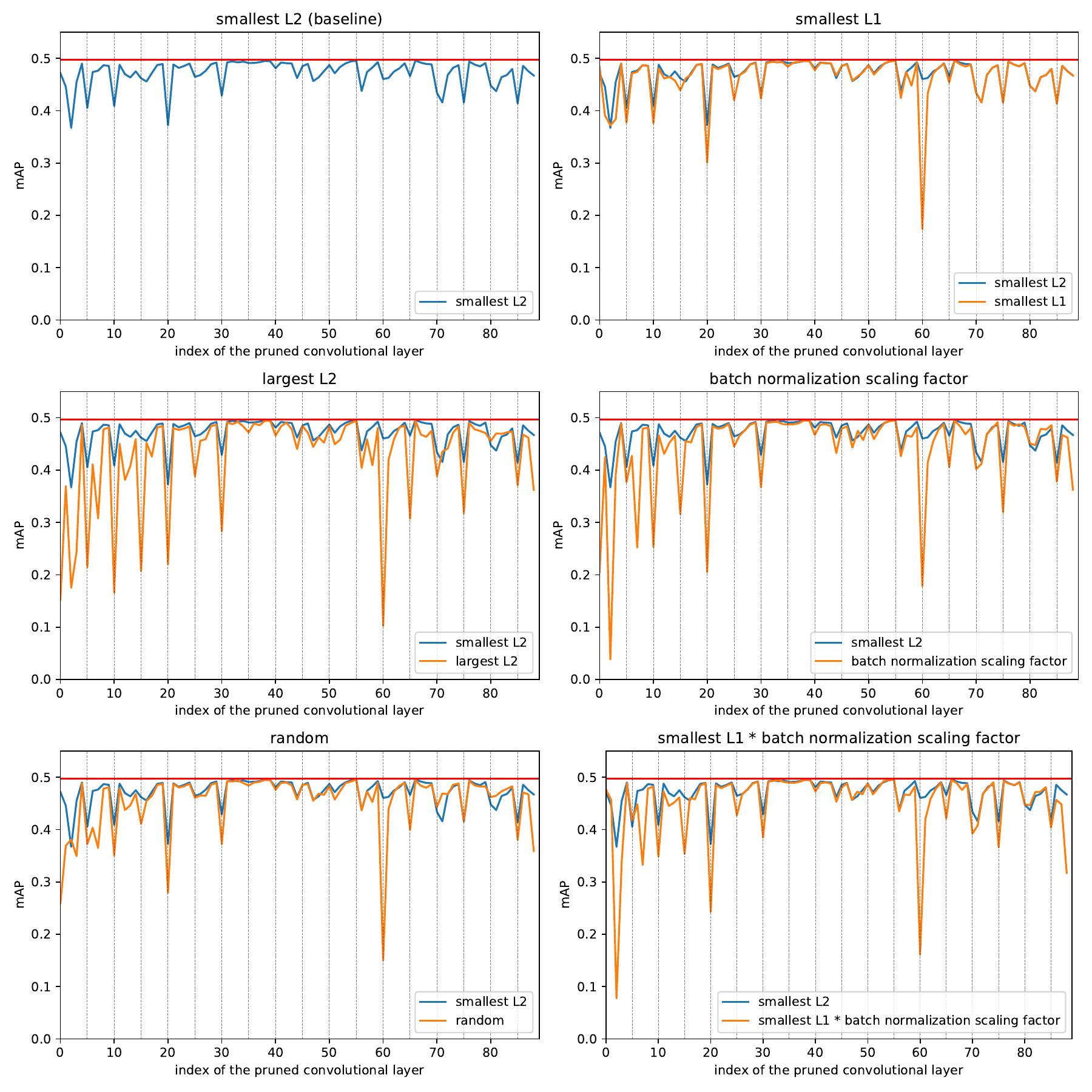}
  \caption{Different criteria (orange) vs. smallest L2 (blue).}
  \label{fig:filter_criteria_sensitivity}
\end{figure}

%Now, after we found out that we need to prune filters after fusing the batch normalization layers into the convolutional layers, we compare different criteria for pruning filters that have been used in other works. 
Comparison of different criteria for filter pruning using sensitivity analysis (see Table~\ref{tab:average_mAP} and Figure~\ref{fig:filter_criteria_sensitivity}) has revealed, that pruning filters with the smallest L2-norm works the best for \texttt{YOLOv7} and especially outperforms pruning large or random filters. The difference is especially evident in the earlier layers where lower-level features are extracted, whereas deeper layers of the backbone (layers 31 to 40) are more resistant to pruning. Selecting filters with the smallest L2-norm works marginally better than the L1-norm for most layers. The criterion proposed by Liu et al.~\cite{liu2017learning} of selecting filters based on the batch normalization layer's scale factor as well as multiplication with the L1-norm~\cite{zhang2021pruned} also performed worse than the L2-norm criterion. %The reason for this likely is that our pre-trained YOLOv7 model did not impose regularization on the batch normalization scale factor, unlike the aforementioned methods. The weight decay (L2-regularization) that was applied during the training of YOLOv7 can be seen as a form of sparsity training which is probably the reason why the L2-norm criterion works the best for us. This might also explain why pruning after fusing the batch normalization layer performs worse. The weights of the convolutional layers get mixed with the parameters of the batch normalization layers which had no regularization imposed on them. This diminishes the ability to select unimportant filters based on the L2-norm after the fusion of the batch normalization layers.

\newpage
\clearpage

\begin{figure*}[tb]
  \centering
  \begin{subfigure}[t]{\textwidth}
    \centering
    \includegraphics[width=\textwidth]{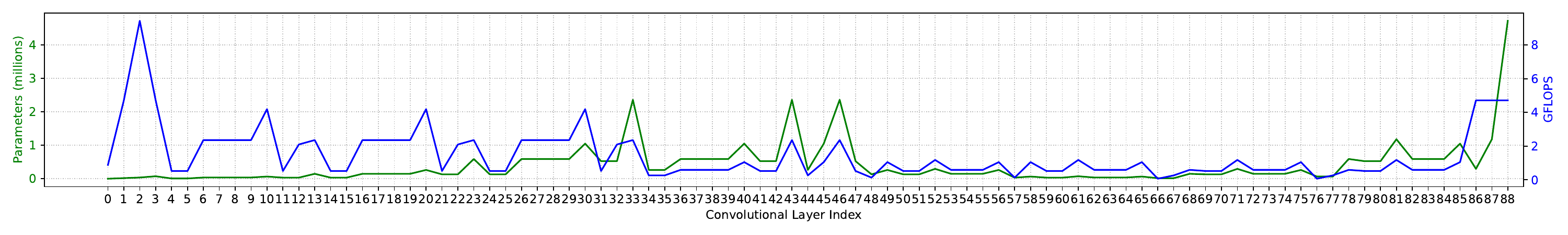}
    \caption{\texttt{YOLOv7}: distribution of parameters and FLOPs per layer}
  \end{subfigure}
  \begin{subfigure}[t]{\textwidth}
    \centering
    \includegraphics[width=\textwidth]{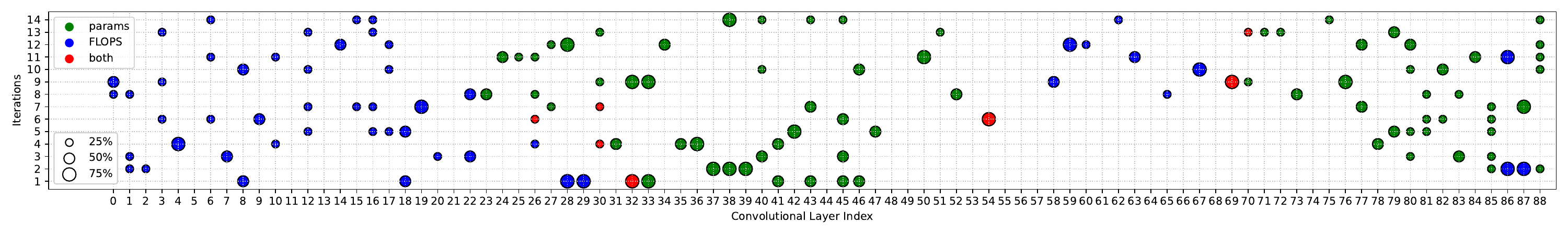}
    \caption{\texttt{YOLOv7}: filters pruned per layer and iteration. Blob size represents the pruning rate.}
  \end{subfigure}
    \begin{subfigure}[t]{\textwidth}
    \centering
    \includegraphics[width=\textwidth]{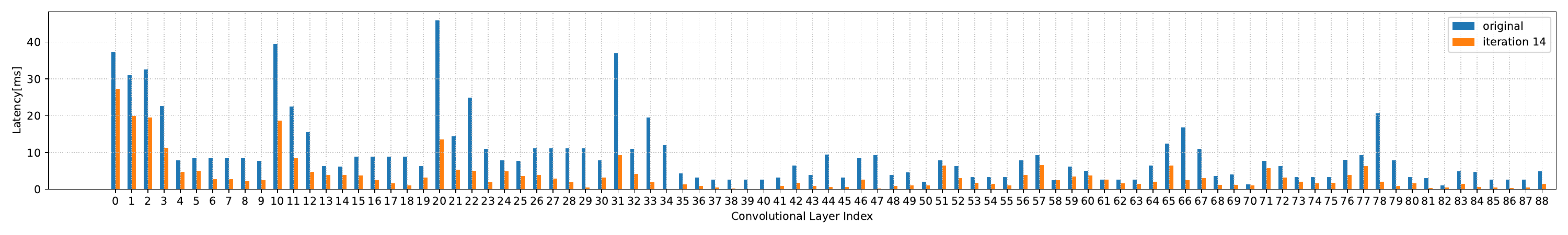}
    \caption{\texttt{YOLOv7}: latency per layer, evaluated on FPGA}
  \end{subfigure}
    \begin{subfigure}[t]{\textwidth}
    \centering
    \includegraphics[width=\textwidth]{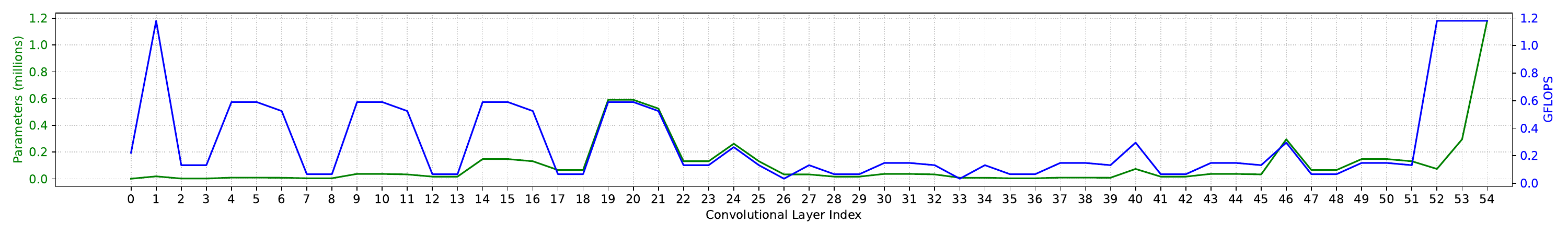}
     \caption{\texttt{YOLOv7-tiny}: distribution of parameters and FLOPs per layer}
  \end{subfigure}
  \begin{subfigure}[t]{\textwidth}
    \centering
    \includegraphics[width=\textwidth]{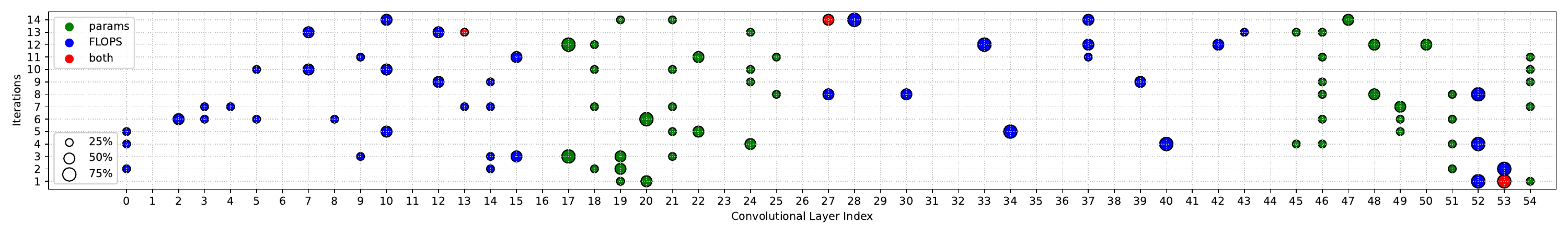}
    \caption{\texttt{YOLOv7-tiny}: filters pruned per layer and iteration. Blob size represents the pruning rate.}
  \end{subfigure}
  \begin{subfigure}[t]{\textwidth}
    \centering
    \includegraphics[width=\textwidth]{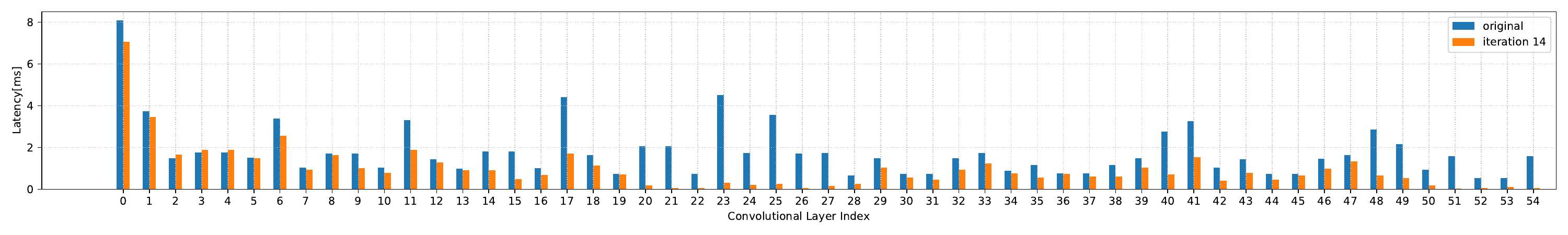}
    \caption{\texttt{YOLOv7-tiny}: latency per layer, evaluated on FPGA}
  \end{subfigure}
  \caption{Analysis of the distribution of parameters and FLOPs over layers in unpruned \texttt{YOLOv7} and \texttt{YOLOv7-tiny} models (a, d); pruned parameters and FLOPs and the corresponding pruning rates, selected by our algorithm (b, e); latency per layer for the models deployed to FPGA (c, f).}
  \label{fig:pruning-per-layer}
\end{figure*}

\newpage
\clearpage

\subsection{Layer Selection and Performance}
\begin{figure}[h]
  \centering
  \begin{subfigure}[t]{0.455\columnwidth}
    \centering
    \includegraphics[width=\textwidth]{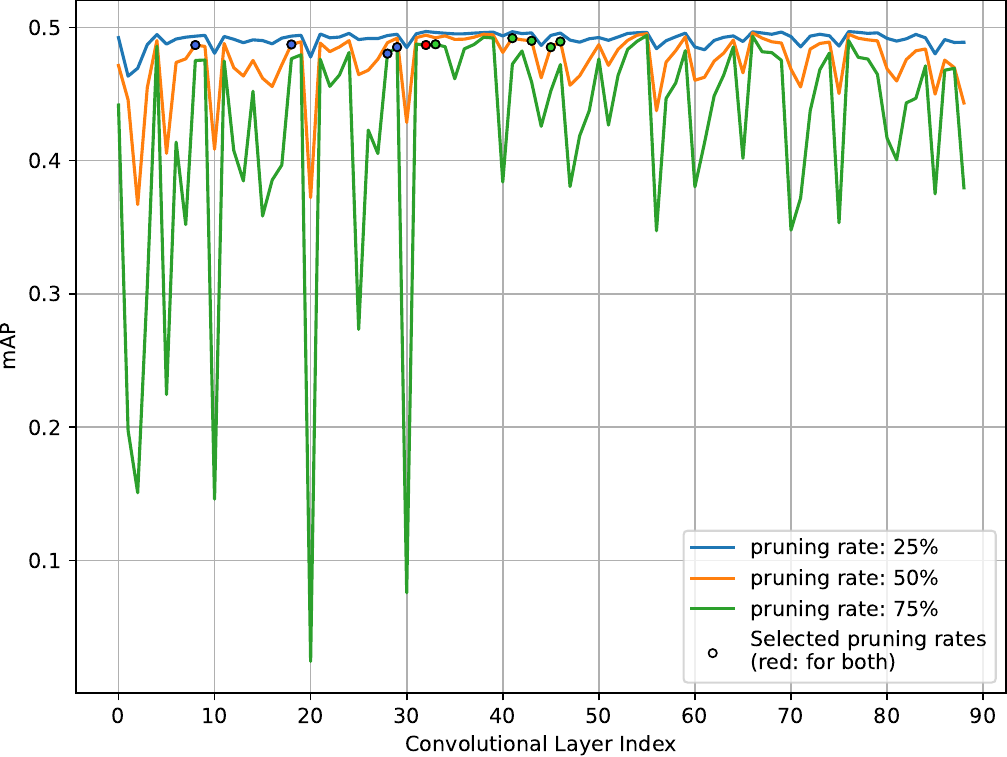}
    \label{fig:yolo_sa_map}
    \caption{\texttt{YOLOv7}: mAP}
    \end{subfigure}
 \begin{subfigure}[t]{0.455\columnwidth}
    \centering
    \includegraphics[width=\textwidth]{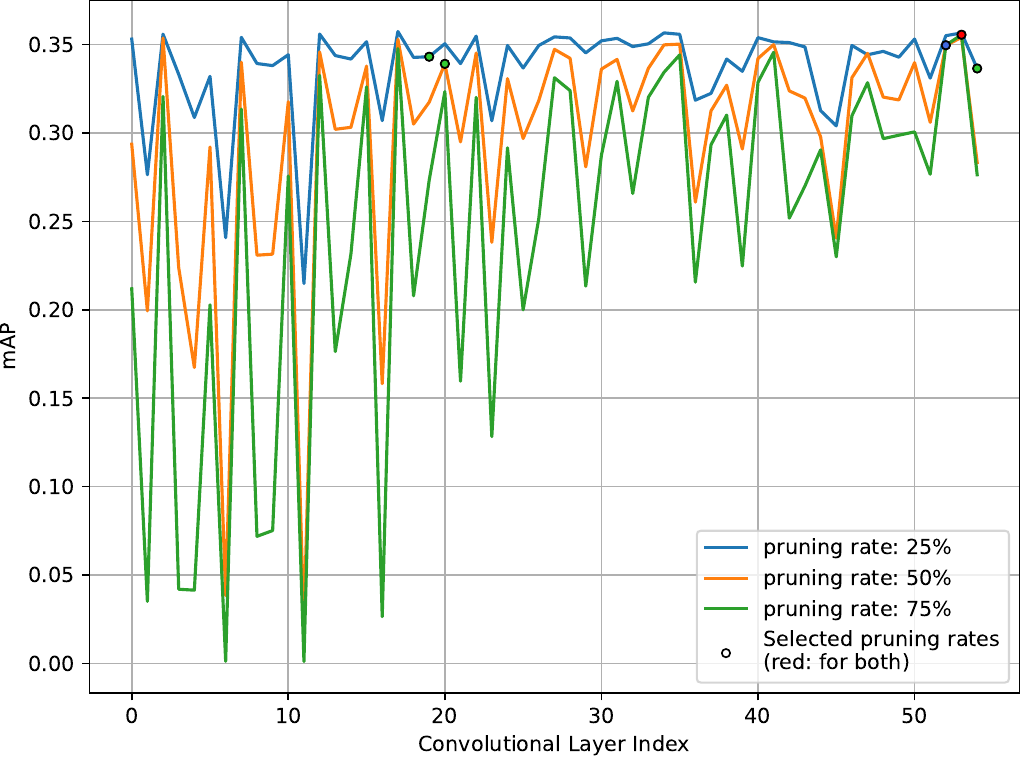}
    \label{fig:tiny_sa_map}
    \caption{\texttt{YOLOv7-tiny}: mAP}
  \end{subfigure}
  
\begin{subfigure}[t]{0.455\columnwidth}
    \centering
    \includegraphics[width=\textwidth]{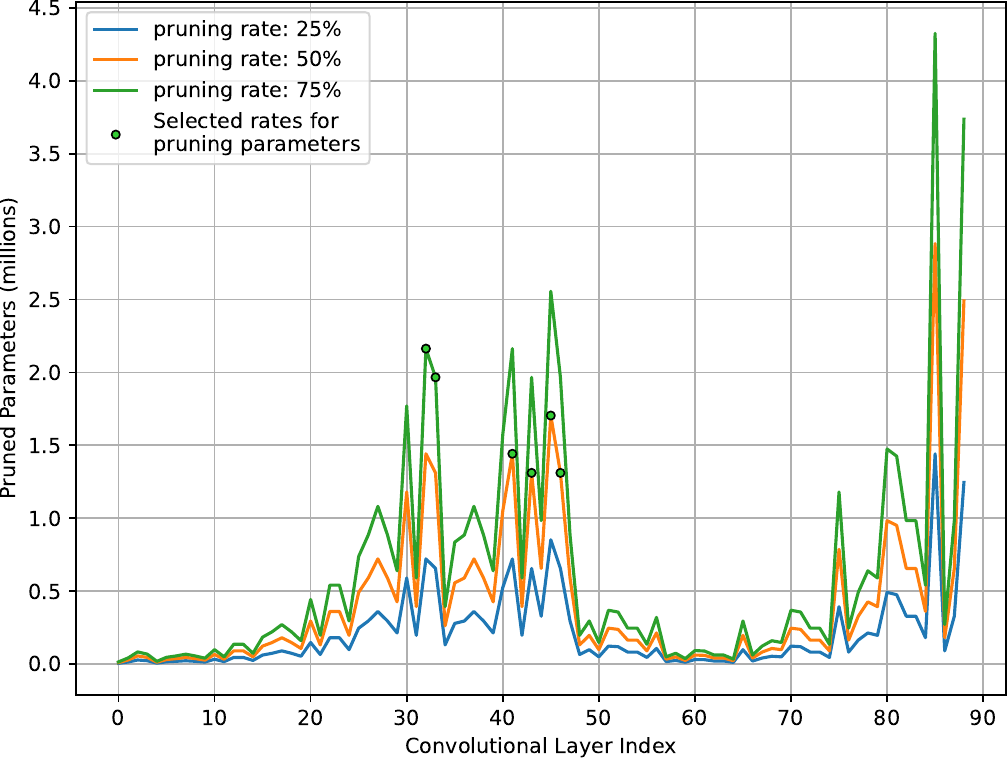}
    \label{fig:yolo_sa_parameters}
    \caption{\texttt{YOLOv7}: parameters}
  \end{subfigure}
  \begin{subfigure}[t]{0.455\columnwidth}
    \centering
    \includegraphics[width=\textwidth]{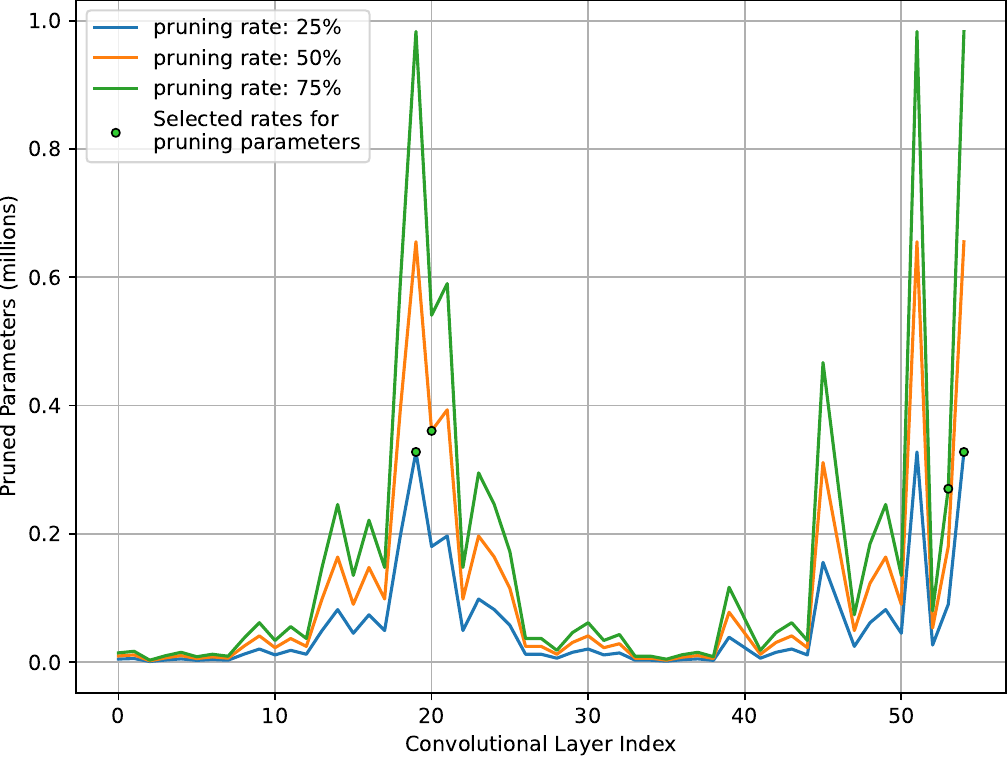}
    \label{fig:tiny_sa_parameters}
    \caption{\texttt{YOLOv7-tiny}: parameters}
  \end{subfigure}
  
\begin{subfigure}[t]{0.455\columnwidth}
    \centering
    \includegraphics[width=\textwidth]{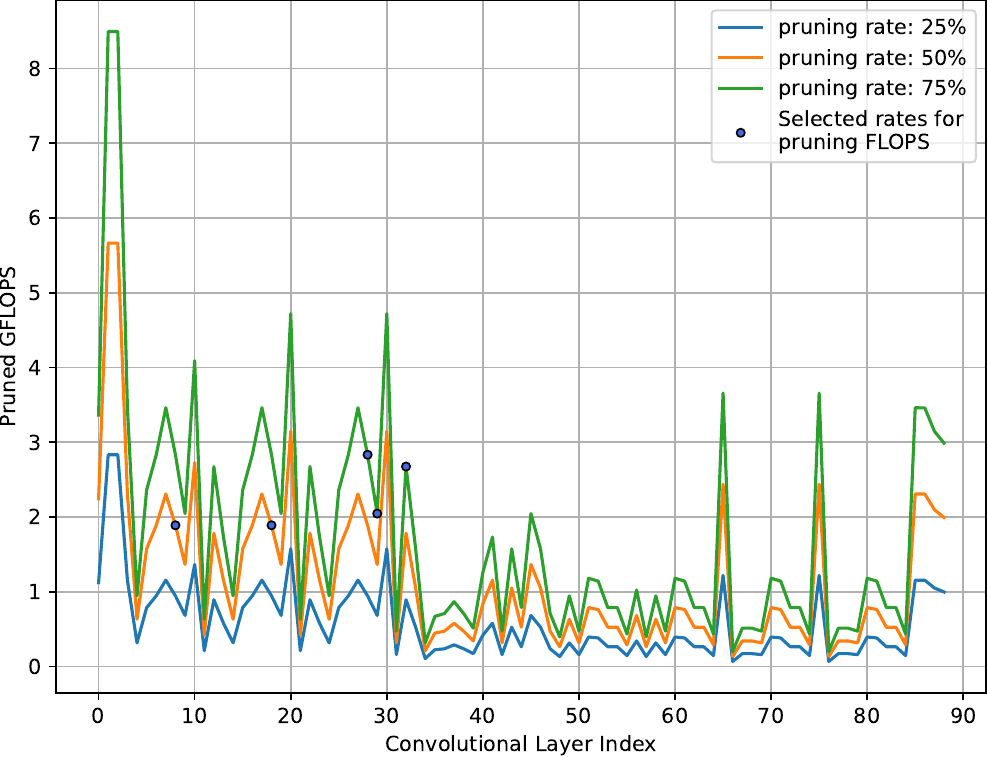}
    \label{fig:yolo_sa_FLOPs}
    \caption{\texttt{YOLOv7}: FLOPs}
  \end{subfigure}
  \begin{subfigure}[t]{0.455\columnwidth}
    \centering
    \includegraphics[width=\textwidth]{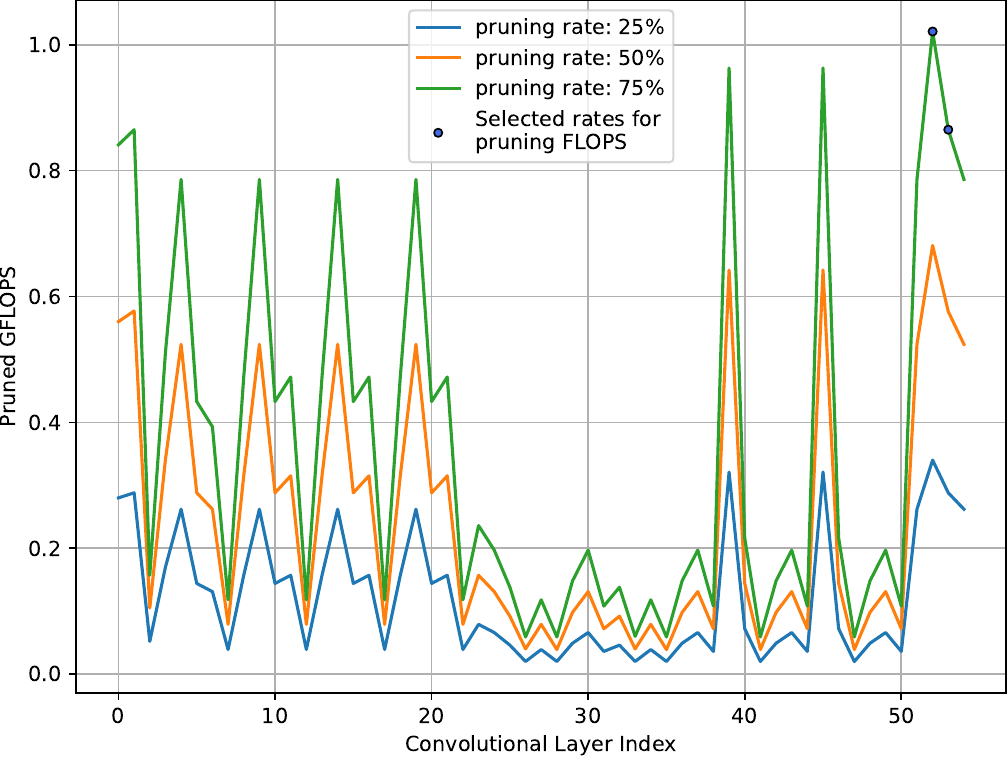}
    \label{fig:tiny_sa_FLOPs}
    \caption{\texttt{YOLOv7-tiny}: FLOPs}
  \end{subfigure}
  \caption{Sensitivity analysis of the models for the first pruning iteration.}
  \label{fig:sensitivity-analysis}
\end{figure}

We exemplary show the process of layer and pruning rate selection in Figure~\ref{fig:sensitivity-analysis} for the first pruning iteration and the selected pruning rates. Analysis of layers selected for pruning across iterations (see Figure~\ref{fig:pruning-per-layer}, b and e) shows distinct pruning patterns. First, early layers in both models are pruned based on FLOPs, which aligns with the distribution of FLOPs per layer (cf. Figure~\ref{fig:pruning-per-layer}, a and d). This applies to a lesser extent to the distribution of parameters. For both architectures, most FLOPs are in the backbone layers (layers 0 to 40 for \texttt{YOLOv7}, layers 0 to 28 for \texttt{YOLOv7-tiny}), and our method also selects FLOPs-based pruning for early backbone layers. Also, larger pruning rates are used for earlier layers which contain larger feature maps and correspondingly more FLOPs. Additionally, there is a rise in some parameters and FLOPs in the last layers - here also large pruning rates are selected.

Certain layers have been repeatedly selected for pruning, whereas others were never selected, thus demonstrating robustness to pruning. This effect is especially pronounced in the \texttt{YOLOv7-tiny} architecture and suggests that this model is significantly less over-parameterized. Also, smaller pruning rates were selected for \texttt{YOLOv7-tiny}.

\begin{figure}[t]
  \centering

  \begin{subfigure}[t]{\columnwidth}
    \centering
    \includegraphics[width=\textwidth]{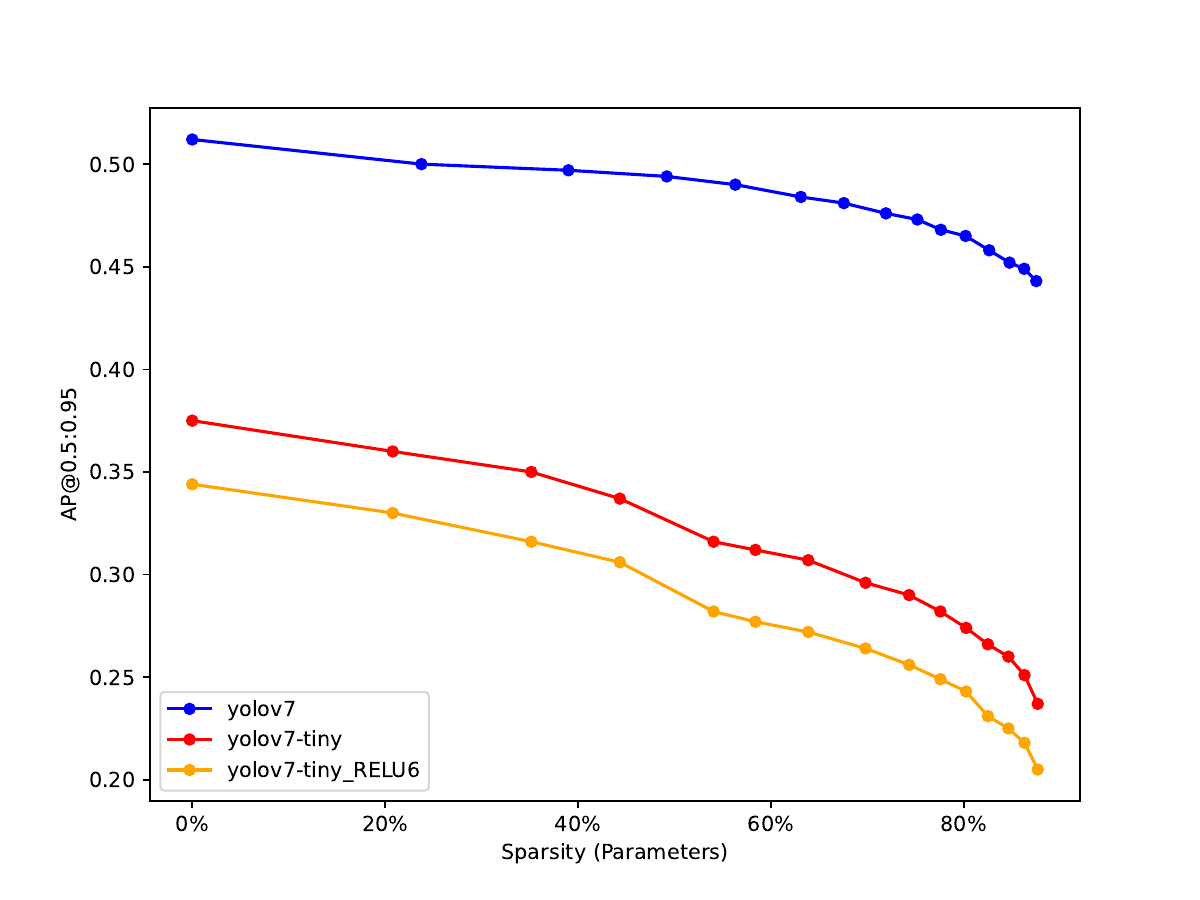}
    %\caption{AP vs. Parameters}
  \end{subfigure}
  
  \begin{subfigure}[t]{\columnwidth}
    \centering
    \includegraphics[width=\textwidth]{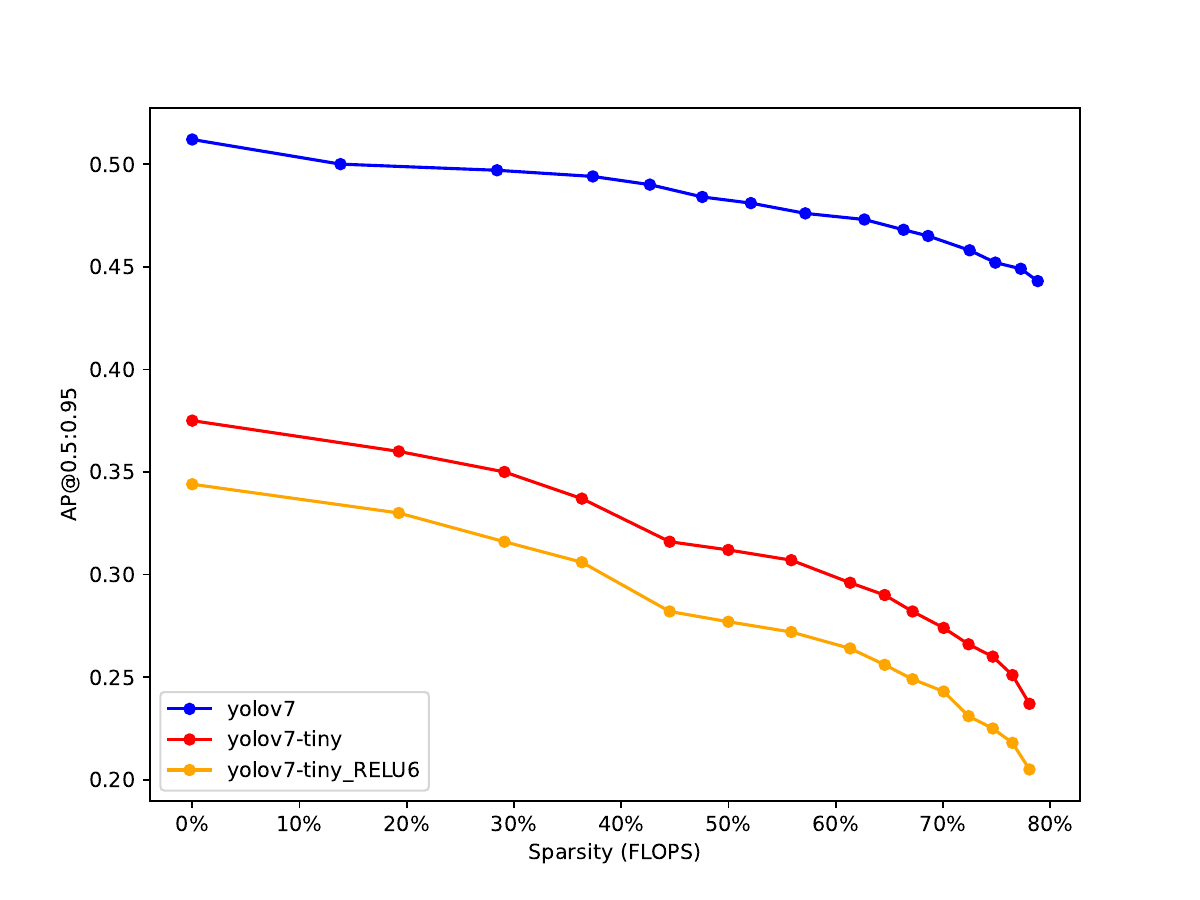}
    %\caption{AP vs. FLOPs}
  \end{subfigure}
  \caption{AP for different sparsity levels, measured either in terms of the number of parameters or FLOPs.}
  \label{fig:ap-vs-params-flops}
\end{figure}

Furthermore, we have analyzed how accuracy diminished with the growing sparsity (see Figure~\ref{fig:ap-vs-params-flops}). For both architectures, the three first iterations have brought the most gain in sparsity, both in the number of parameters and FLOPs. Starting from iteration four, the accuracy starts to degrade. The last iterations bring only a small improvement in sparsity, showing that the compression potential has been exhausted.

\subsection{Evaluation on Hardware}

To assess the effect of pruning, we evaluate two different edge hardware setups. Notably, model sparsity does not necessarily result in an inference speedup. Furthermore, most work report gain in acceleration and memory consumption for the whole model compression pipeline, which usually involves further methods apart from pruning, In particular, the effect of quantization is usually measured together with pruning. To realistically assess the gain achieved only via the application of the proposed pruning, we compare inference latency for an unpruned model and across pruning iterations.

For evaluation of the inference speed of the pruned models on embedded hardware, we used the TVM deployment framework\footnote{https://tvm.apache.org} for the Gemmini accelerator~\cite{gemmini-dac}\footnote{https://github.com/ucb-bar/gemmini}. Our pipeline is based on the one presented by Peccia et al.~\cite{peccia2022integration}. 

We implemented a system-on-chip architecture working at 150 MHz composed of a RocketChip core, a 32x32 size Gemmini, and a 2048 kB L2 cache on a Xilinx ZCU102 FPGA. For comparison against commercially available embedded hardware, we deployed the same models using the TVM Deep Learning framework on an NVIDIA Jetson AGX Xavier. Within the used pipeline, the PyTorch model is exported via ONNX to Tensorflow. The model is then quantized to int8 using Tensorflow Lite, translated to TVM, and further integrated into Gemmini via microTVM.

Because the SiLu activation function, used in \texttt{YOLOv7-tiny} is not supported by Gemmini, we have replaced it with RELU6. Furthermore, we have reduced the resolution to $480\times480$ for faster inference. This model is referred to as \texttt{YOLOv7-tiny\_RELU6}. The mAP before pruning has thus dropped from 0.360 to 0.339 after resolution reduction and to 0.331 after replacing the activation function (cf. Figure~\ref{fig:ap-vs-params-flops}). % The resulting \texttt{YOLOv7-tiny\_RELU6} model was then pruned and deployed to hardware. 

Furthermore, the overview of latency gains per layer (see Figure~\ref{fig:pruning-per-layer}) demonstrates speedup for each layer.  Layers, which were not deliberately pruned, profit from the reduced feature maps as a result of pruning of preceding layers (e.g., layers 48 and 59 in \texttt{YOLOv7}, layers 29 and 31 in \texttt{YOLOv7-tiny}). Latency peaks mostly correspond to layers with a large number of FLOPs, which were sometimes pruned moderately or not pruned at all (e.g., layer 20 in \texttt{YOLOv7}, layer 11 in \texttt{YOLOv7-tiny}). This suggests that the layer selection algorithm can be further improved.

\begin{figure}[h]
  \centering
    \begin{subfigure}[t]{\columnwidth}
    \centering
    \includegraphics[width=\textwidth]{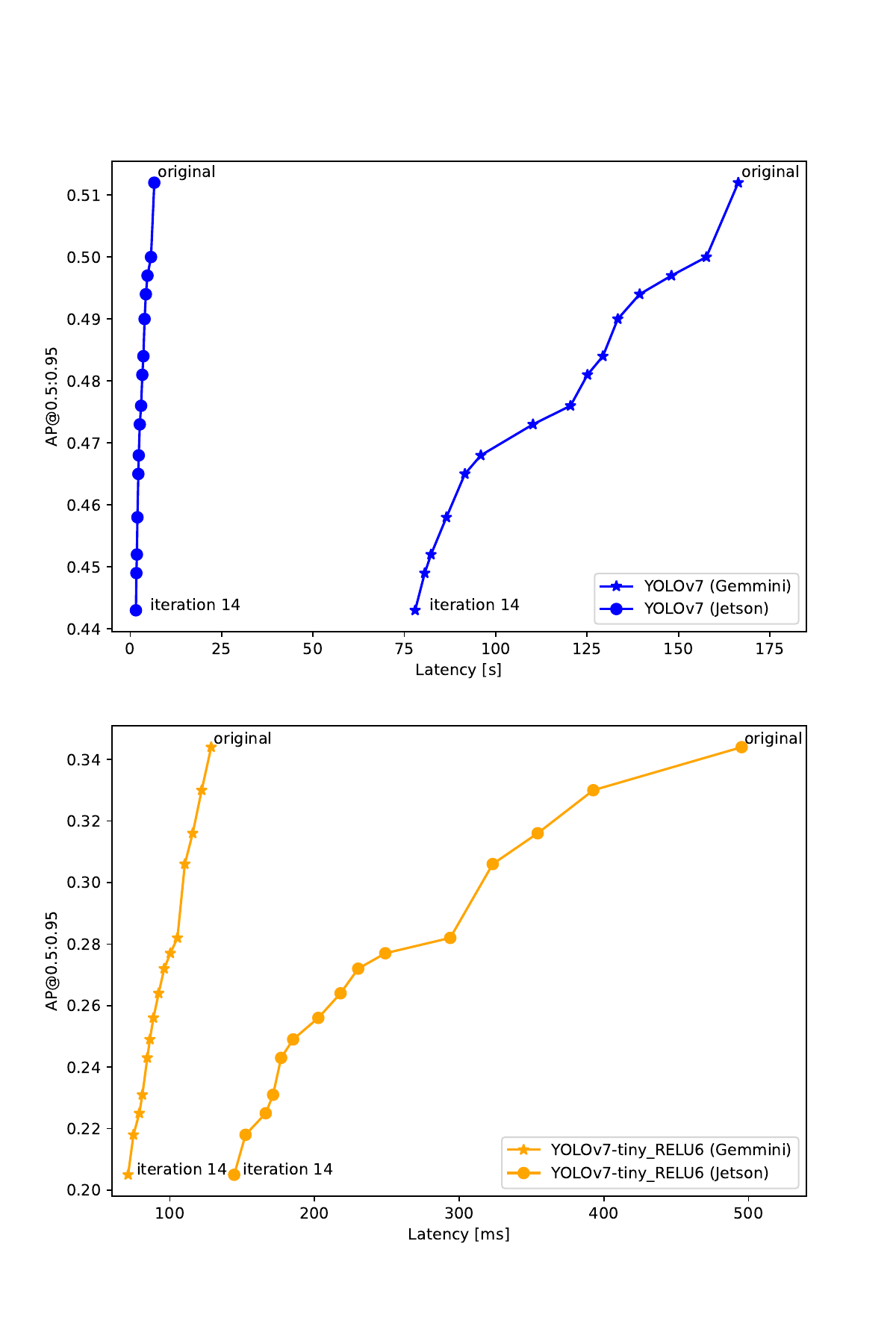}
    %\caption{AP vs. speed}
  \end{subfigure}
  \caption{Performance of YOLOv7 and YOLOv7-tiny for different sparsity levels.}
  \label{fig:hardware-speed}
\end{figure}

Figure~\ref{fig:hardware-speed} demonstrates latency over the pruning iterations. Similar to model sparsity, we observe that major improvements are made during the first iterations, especially for \texttt{YOLOv7-tiny\_RELU6}.  Starting from iteration five, AP drops without a significant improvement in latency. The proposed iterative approach can thus, among others, be used to select a model for the deployment based on the trade-off between accuracy and inference speed. Interestingly,  the smallest pruned YOLOv7 model that we obtained has less parameters but is more accurate than YOLOv7-tiny.

Comparing the performance of the pruned models on different hardware, we could observe that the obtained gain in speed for our FPGA implementation reaches  45\% for \texttt{YOLOv7} and 37\% for \texttt{YOLOv7-tiny}, whereas the speedup for convolutional layers only is 54\% and 40\% correspondingly. For the models deployed to Jetson,  the overall speedup is 66\% for \texttt{YOLOv7} and 64\% for \texttt{YOLOv7-tiny}, whereas for convolutional layers it is 66\% and 65\% correspondingly. Using the proposed pruning approach, we could accelerate inference of \texttt{YOLOv7-tiny} on Jetson from 4.02 to 6.92 FPS and on FPGA from 7.78 to 14.06 FPS, thus reaching real-time performance.

\section{Conclusion}
In this work, we have proposed an automated iterative sensitivity-based filter pruning for object detectors with concatenation-based architectures. Our approach relies on a layer connectivity graph, which helps to correctly handle concatenation layers, which are usually overseen in current works. Using sensitivity analysis, our approach automatically selects layers for pruning as well as pruning rates independently for each iteration. After each iteration, fine-tuning is applied to restore accuracy.

The proposed approach was exemplary and evaluated on \texttt{YOLOv7} and \texttt{YOLOv7-tiny} architectures on the COCO dataset and deployed to FPGA and NVIDIA Jetson AGX Xavier. For pruned \texttt{YOLOv7-tiny}, we could reach a real-time inference speed of 14 FPS on FPGA. By analyzing the trade-off between accuracy and performance, the most suitable model for deployment can be selected across iterations. 

Finally, the proposed approach can as well be applied to any neural network architecture containing concatenation layers. In particular, direct application to further YOLO versions is possible.

\section*{Acknowledgement}

This research is funded by the German Federal Ministry of Education and Research within the project "GreenEdge-FuE“, funding no. 16ME0517K.

{\small
\bibliographystyle{IEEEtran}
\bibliography{references}
}

\end{document}